\documentclass[sigconf]{acmart}

\usepackage{graphicx, caption, subcaption}
\usepackage{xspace}
\usepackage{makecell}

\usepackage{booktabs}
\usepackage{multirow}
\usepackage{float}
\usepackage{adjustbox}
\usepackage{subcaption}
\newcommand{\bfsection}[1]{\vspace*{0.1cm}\noindent\textbf{#1.}}

\newcommand{\bfsections}[1]{\vspace*{0.1cm}\textbf{#1.}}

\newcommand{\etal}{\textit{et al}.}
\newcommand{\ie}{\textit{i}.\textit{e}. }
\newcommand{\eg}{\textit{e}.\textit{g}. }
\def\etc{\emph{etc}. }

\usepackage{enumitem}

\AtBeginDocument{%
  \providecommand\BibTeX{{%
    \normalfont B\kern-0.5em{\scshape i\kern-0.25em b}\kern-0.8em\TeX}}}

\copyrightyear{2022}
\acmYear{2022}
\setcopyright{rightsretained}

\acmConference[SIGSPATIAL '22]{The 30th International Conference on Advances in Geographic Information Systems}{November 1--4, 2022}{Seattle, WA, USA}

\acmBooktitle{The 30th International Conference on Advances in Geographic Information Systems (SIGSPATIAL '22), November 1--4, 2022, Seattle, WA, USA}
\acmDOI{10.1145/3557915.3560946} \acmISBN{978-1-4503-9529-8/22/11}

\acmSubmissionID{39}

\usepackage[symbol]{footmisc}

\begin{document}

\title{EgoSpeed-Net: Forecasting Speed-Control in Driver Behavior from Egocentric Video Data}
\renewcommand{\shorttitle}{EgoSpeed-Net: Forecasting Driver Behavior}

\author{Yichen Ding$^a$, Ziming Zhang$^b$,  Yanhua Li$^b$, Xun Zhou$^{a*}$}
\affiliation{%
  \institution{$^a$University of Iowa, Iowa, $^b$Worcester Polytechnic Institute, Massachusetts}
  }
 
\email{{yichen-ding,xun-zhou}@uiowa.edu, {zzhang15,yli15}@wpi.edu}

\renewcommand{\shortauthors}{Ding et al.}

\begin{abstract}
  Speed-control forecasting, a challenging problem in driver behavior analysis, aims to predict the {\em future} actions of a driver in controlling vehicle speed such as braking or acceleration. In this paper, we try to address this challenge solely using egocentric video data, in contrast to the majority of works in the literature using either third-person view data or extra vehicle sensor data such as GPS, or both.To this end, we propose a novel graph convolutional network (GCN) based network, namely, {\bf EgoSpeed-Net}. We are motivated by the fact that the position changes of objects over time can provide us very useful clues for forecasting the speed change in future. We first model the spatial relations among the objects from each class, frame by frame, using fully-connected graphs, on top of which GCNs are applied for feature extraction. Then we utilize a long short-term memory network to fuse such features per class over time into a vector, concatenate such vectors and forecast a speed-control action using a multilayer perceptron classifier. We conduct extensive experiments on the Honda Research Institute Driving Dataset, and demonstrate superior performance of EgoSpeed-Net. 
\end{abstract}



\maketitle

\footnotetext{\hspace{-15pt}$^*$ Xun Zhou is the corresponding author.}

\section{Introduction}
\label{sec:intro}

Understanding and predicting driver behavior is an important problem for road safety, transportation, and autonomous driving. The National Highway Transportation Safety Administration claims that 94\%-96\% of auto accidents are due to different types of human errors and the majority of them belong to driver negligence or carelessness~\cite{nhtsa2017}. Shinar \etal.~\cite{shinar1998automatic} also suggest that drivers are less aware of their skills with more experience, as most of the actions taken by the driver during the driving are unconscious. Building accurate prediction models helps us better understand how and why drivers make their decisions. Such knowledge can also be incorporated into the design of autonomous vehicles and driving simulators.

\begin{figure}[t!]
\centering
\begin{subfigure}{0.49\textwidth}
    \includegraphics[width=\textwidth]{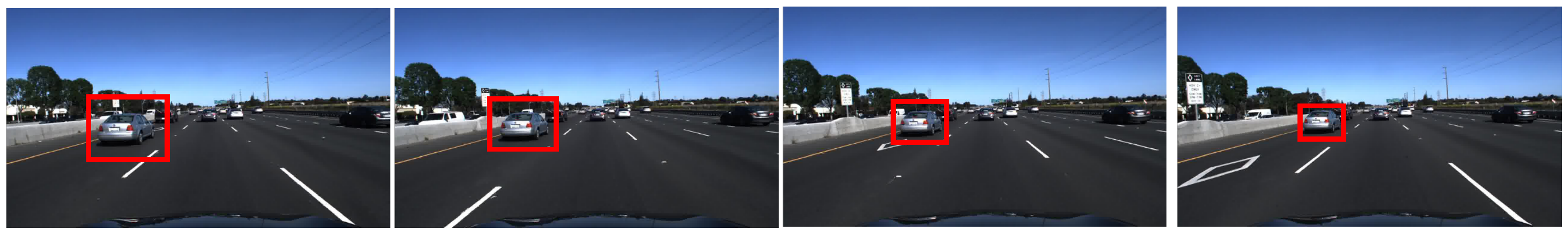}
    \vspace{-0.15in}
    \caption{Highway}
    \label{fig:intro-accel-highway}
\end{subfigure}
\hfill
\begin{subfigure}{0.49\textwidth}
    \includegraphics[width=\textwidth]{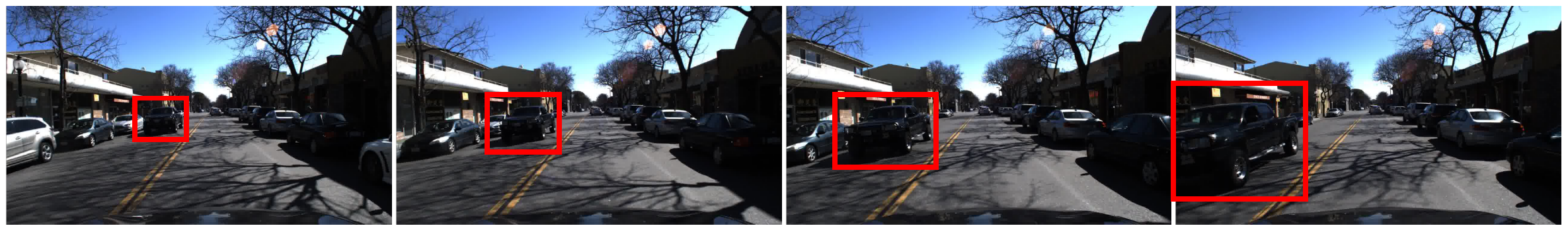}
    \vspace{-0.15in}
    \caption{Urban}
    \label{fig:intro-accel-urban}
\end{subfigure}
\vspace{-0.1in}
\caption{Illustration of a typical challenge in forecasting speed-control from egocentric video data, where the same acceleration action is taken in highway and urban scenarios, respectively, but leading to contradictorily visual observations of the same vehicles.}
\label{fig:intro-accel}
\vspace{-0.2in}
\end{figure}

In this paper we aim to address a challenging problem in driver behavior analysis, namely, speed-control forecasting, using only egocentric (\ie first-person view point) videos collected along the trip. {Instead of simply predicting the speed or trajectory of a vehicle, we focus on the prediction of acceleration and braking behaviors of drivers} as they are the most common and essential actions. In addition, a large number of road fatalities are related to speed-control issues~\cite{nhtsa2019}. In particular, we formulate the problem using multi-class prediction: given a sequence of egocentric video frames, we try to predict the speed-control action (\ie slight acceleration, full acceleration, slight braking, full braking) in the next (few) frame. {Note that we use solely egocentric video data as the input because it best represents the driver’s first-person visual input during driving. 
We use driver’s braking or accelerating information from pedal sensor data to define the ground-truth behaviors. 
We do not use the vehicle’s acceleration or GPS information in the input because they are not part of the road environment and may not be available to the driver in real-time. They are therefore unlikely to have a major and immediate impact on drivers’ speed-control behaviors.} 

\bfsection{Challenges}
To address the problem of speed-control forecasting using egocentric video data, following human driving experience, we are motivated by the fact that the position changes of objects over time can provide us very useful clues for forecasting the speed change in future. However, in practice we are facing the following challenges:
\begin{itemize}[nosep, leftmargin=*]
    
    \item The same speed-control action in different scenarios may lead to different visual observations that could confuse the prediction model significantly. As shown in Figure~\ref{fig:intro-accel-highway}, even the labeled car is moving away from the driver's field of view, his/her speed is still increasing (we know this due to the ground-truth data). Similarly we know that in Figure~\ref{fig:intro-accel-urban} the driver is still speeding up when the labeled car is approaching. Such visual contradictions led by the same speed-control actions widely exist in the real world as well as our evaluation data.
    
    \item Drivers are heavily influenced by their surroundings such as traffic conditions, road conditions, nearby vehicles,~\etc It will be challenging to identify the consistent relationship, spatially and temporally, among such objects/stuff using egocentric videos that are useful for our prediction.
    
    \item The scenes in the egocentric video data are more complicated than those in static cameras because of rapidly dynamical change in objects and background, especially in driving scenarios.
    
\end{itemize}

\bfsection{Approaches}
Prior works on driver/human behavior prediction include mobility-based methods and vision-based methods. The first group predict behaviors (\eg movement patterns) using GPS trajectories of vehicles or spatial data~\cite{oliver2014significant,yuan2018hetero}. These techniques do not consider visual input. Vision-based methods typically use fixed camera video data to recognize actions, behaviors, or predict moving paths of pedestrians or drivers~\cite{liang2020garden,shafiee2021introvert}. Some of these works use RCNN-based methods to extract objects from raw pixels. Others model the relationship between objects in the scene as graphs~\cite{hu2018relation,gkioxari2018detecting}. Few recent works in computer vision consider egocentric video data as the input. However, they either predict the behaviors of objects in the scene (not the observer), or use other auxiliary data to help the learning (\eg spatial trajectories, vehicle sensor readings)~\cite{qiu2021egocentric,qiu2021indoor}.

In contrast, our problem differs from these prior works. We aim to predict the future speed-control behavior of a driver, rather than objects in the scene. 
Note that, the camera collecting our data is also moving, while in the literature the majority of the methods use videos from fixed cameras even from the first-person view. 
To address the limitations of prior works, we propose an EgoSpeed-Net to learn the spatiotemporal relations among the objects in the videos.  

\bfsection{Contributions}
Our contributions are listed as follows:
\begin{itemize}[nosep, leftmargin=*]

\item To the best of our knowledge, we are the {\em first} to address the problem of speed-control forecast for driver behavior analysis purely using egocentric video data. 
    
\item We propose a novel deep learning solution, namely, EgoSpeed-Net, a seamless integration of graph representations of videos, GCNs~\cite{kipf2017semi}, LSTMs, and an MLP that can be trained end-to-end effectively and efficiently. 
    
\item We demonstrate superior performance of our EgoSpeed-Net on the Honda Research Institute Driving Dataset (HDD)~\cite{ramanishka2018toward}, compared with state-of-the-art methods.

\end{itemize}

The rest of this paper is organized as follows. Section \ref{sec:related} discusses related work. In Section \ref{sec:method}, we define the driver behavior prediction problem and elaborate our proposed EgoSpeed-Net framework. Section \ref{sec:evaluation} presents comprehensive evaluation results. Finally, we conclude the paper in Section \ref{sec:conclusion}.

\begin{figure*}[t]
    \centering
    \includegraphics[width=0.99\linewidth]{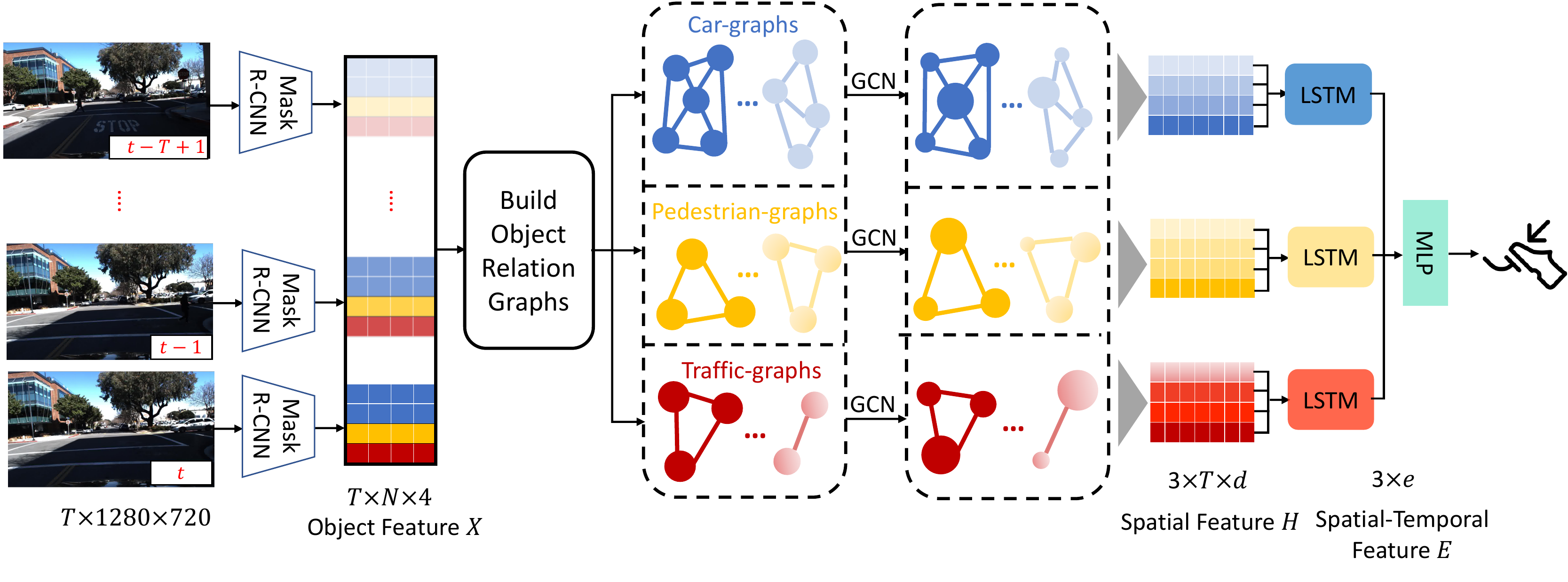}
    \caption{Framework of EgoSpeed-Net for speed-control forecasting.}
    \label{fig:model}
\end{figure*}

\section{Related Work}
\label{sec:related}
\bfsection{Mobility-based behavior analysis}
Existing works mainly focus on pedestrian trajectory prediction~\cite{xu2018encoding,mangalam2020not} and cyclist behavioral studies~\cite{huang2016cyclist,ding2020cycling}. 
For example, Mangalam \etal~\cite{mangalam2020not} proposed PECNet to forecast distant pedestrian trajectories conditioned on their destinations. 
Huang \etal~\cite{huang2016cyclist} represented bicycle movements by taking other road users into consideration using social force model. Few existing works~\cite{liu2014trajectory,altche2017lstm} 
pay attention to driver trajectories prediction. Liu \etal~\cite{liu2014trajectory} estimated the lane change behavior of drivers to predict trajectories. Given a partial GPS trajectory, without any visual input such as images or videos, these methods aim to predict either the destination or the next location along the trajectory. {Recent work considered map information and social context. Zaech \etal~\cite{zaech2020action} predicted action sequence of vehicles in urban environments with the help of HD maps. Nevertheless, we only consider egocentric video as input.}

\bfsection{Vision-based behavior analysis}
In recent years, the applications of deep networks on video analysis have been conducted widely.
\begin{itemize}[nosep, leftmargin=*]
    \item {\em Non-egocentric video data:} Traditionally, most works in driver behavior analysis focus on the detection of driver drowsiness and distraction such as drowsing driving, tailgating, lane weaving behavior, \etc Many driver inattention monitoring systems~\cite{volkswagen1,you2013carsafe} have been designed by taking advantage of eye state analysis~\cite{cyganek2014hybrid,jo2014detecting}, facial expression analysis~\cite{abtahi2011driver,abtahi2013yawning}, driver physiological analysis~\cite{simon2011eeg,baek2011smart}, \etc
    Recent studies took the scene context into consideration to predict the pedestrian trajectory~\cite{kosaraju2019social,sadeghian2019sophie,liang2019peeking} 
    and vehicle movements analysis~\cite{hirayama2012detection,choi2014head}. Liang \etal~\cite{liang2020garden} predicted multiple future trajectories using multi-scale location decoders, while Introvert~\cite{shafiee2021introvert} proposed a conditional 3D visual attention mechanism to infer human-dependent interactions in the scene context. Some of these driver behavior analyses require facial information of the driver themselves, while others focus on video data recorded from fixed cameras. 

    \item {\em Egocentric video data:} Recently, some works have studied on predicting trajectories using egocentric videos~\cite{soo2016egocentric,yagi2018future,yao2019egocentric}. 
    Park \etal~\cite{soo2016egocentric} associated a trajectory with the EgoRetinal map to predict a set of plausible trajectories of the camera wearer, while Yagi \etal~\cite{yagi2018future} inferred future locations of people in the video with their past poses, locations and scales, as well as ego-motion. 
    Qiu \etal~\cite{qiu2021indoor} proposed an LSTM-based encoder-decoder framework for trajectory prediction of human captured in the scene, while Qiu \etal~\cite{qiu2021egocentric} integrated a cascaded cross-attention mechanism in a Transformer-based encoder-decoder framework to forecast the trajectory of the camera wearer. {Meanwhile, some works focus on ego action anticipation~\cite{fernando2021anticipating,grauman2021ego4d}. 
    Fernando \etal~\cite{fernando2021anticipating} anticipated human actions by correlating past with the future with Jaccard similarity measures, while Tran \etal~\cite{tran2021knowledge} distilled information from the recognition model to supervise the training of the anticipation model using unlabeled data. However, most drivers behavior depend on movements of objects they have seen. At current stage, we assume all drivers’ behaviors are taken only based on past videos without any future anticipation.}

\end{itemize}

In contrast, we aim at predicting the behaviors of a driver (\ie the observer) whose movement patterns are likely to be different from those of the pedestrians/vehicles in the scene or the camera wearer. Note that some works utilized multiple modalities. However, the GPS signals are not always accessible or reliable. Our approach only uses the egocentric data as input without any other datasets such as drivers’ trajectories data or vehicles’ various sensor data.

\bfsection{Graph neural networks} For video understanding, early works~\cite{simonyan2014two,wang2016temporal} 
considered the spatial and temporal relationship separately and then fuse the extracted features. Recently, more works pay attention to visual relationships among object instances in space and time~\cite{hu2018relation,gkioxari2018detecting}. 
Inspired by object graphs~\cite{jain2016structural,yuan2017temporal} and GCNs, Wang and Gupta \cite{wang2018videos} proposed to represent videos as space-time region graphs, which model spatial-temporal relations and capture the interactions between human and nearby objects. Xu \etal~\cite{xu2020g} adopted GCNs to localize temporal action with multi-level semantic context in video understanding. Li \etal~\cite{li2021representing} designed MUSLE to produce 3D bounding boxes in each video clip and represent videos as discriminative sub-graphs for action recognition problem. 
  
In contrast, our approach is also based on GCNs, but we differ from them in: 1) We aim to predict the behavior of the observer rather than actors in the scene; 2) The camera collecting our data is always moving, while the majority of the existing methods use videos from fixed cameras; 3) Our approach is designed with multi-view object relation graphs, which can reveal different underlying patterns of moving objects from multiple views of the scene.

\section{Our Approach: EgoSpeed-Net}
\label{sec:method}

Given a sequence of frames from an egocentric video, our goal is to forecast the speed-control action of the driver in the next (few) frame\footnote{By default, for simplicity in explanation we consider the next frame in the paper without explicitly mentioning.}. We denote a video clip at the current time $t$ as $\mathcal{I}_t=\{I^i\}_{i\in[t-T+1, t]}$, where each $I^i$ is an image of frame $i$. We aim at learning a prediction function $f: \mathcal{I}_t \rightarrow y^{t+1}$ to forecast the speed-control action in the next $(t+1)$-frame, $y^{t+1}$, based on the current data $\mathcal{I}_t$, where $y$ is a multi-valued categorical variable (e.g., different levels of acceleration and braking).

\subsection{Overview of EgoSpeed-Net}
The framework of our proposed model is visualized in Figure~\ref{fig:model}. Given a sequence of $T$ frames from the egocentric video data, we first extract feature vectors of objects in the scene, which adopts Mask R-CNN \cite{He_2017_ICCV} to detect objects for each frame. Then, we use the bounding box coordinates to represent objects and select top $N$ objects in each frame according to their identified confidence scores and traffic-related categories (\eg car, pedestrian, traffic light, stop sign, etc.). Specifically, we identify objects and select top $n_{\textit{car}}, n_{\textit{pedestrian}}$, and $ n_{\textit{traffic}}$ for vehicle-category (including cars, buses and trucks), pedestrian-category, and traffic-category (including traffic lights and stop signs)\footnote{Empirically we observe that such six semantic object categories are sufficient for analyzing the driving scenarios in our data.}, respectively. Therefore, the feature dimension is $T \times N \times 4$ after identifying objects, where $N= n_{\textit{car}}+n_{\textit{pedestrian}}+n_{\textit{traffic}}$. 
Formally, the object feature is defined as follows:
\begin{equation}
\begin{split}
    \mathcal{X} & =\{\mathrm{MaskRCNN}(I^{t-T+1}),\dots,\mathrm{MaskRCNN}(I^{t})\} \\
    &  = \{X^{t-T+1},X^{t-T+2},\dots,X^t\},
\end{split}
\end{equation}
where $X^t=\{X^t_j|j \in \{$car, pedestrian, traffic $\} \} \in \mathbb{R}^{N \times 4}$ and $\mathcal{X} \in \mathbb{R}^{T \times N \times 4}$. In detail, for identified objects belonging to category $j$ in frame $t$, we aggregate corresponding feature matrix $X^t_j=\{x^t_{j,i}|i=1,\dots,n_j\} \in \mathbb{R}^{n_j \times 4}$, where $x^t_{j,i} \in \mathbb{R}^4$ is the bounding box coordinates of the $i$-th object in category $j$ in frame $t$, the total number of  top identified objects is $N = \sum n_j$, and $j\in \{$car, pedestrian, traffic$\}$.

Afterwards, based on the bounding box coordinates of the objects, we build object relation graphs, where each node denotes an object. Each edge in the graphs connects two objects from the same category in each frame. Besides, we classify these graphs according to their categories to provide multiple views of the scene.
Next, we perform K-hop localized spectral graph convolution on each object relation graph to capture the local spatial relations among objects in each frame. After graph convolution, we max pool over each object relation graph to aggregate the localized spatial features from the nearby objects, which is in $d$-dimension. For each view, we extract the $T \times d$-dimensional matrix $H$ to represent feature vectors of objects. 

Finally, we apply a standard multi-layer Long Short-Term Memory (LSTM) to learn the temporal dependencies between correlated objects. After graph convolution and a temporal layer, the representations of object relation graphs are fused together to generate spatiotemporal features of objects across space and time. Then, a classifier for predicting the speed-control behavior will be applied on these spatiotemporal features. We adopt a Multi Layer Perceptron (MLP) layer following with a softmax function for the driver behavior classification. 

\subsection{Modeling Spatial Relations among Objects}
As mentioned before, object relation graph is the core of our EgoSpeed-Net. Because the objects identified in each frame are unstructured data and the movements of these objects are dynamic, it is hard to capture these patterns in fixed matrices. Meanwhile, the recent work of Graph Convolution Networks (GCNs)~\cite{kipf2017semi}-based models can successfully learn rich relation information from non-structural data and infer relational reasoning on graph~\cite{wu2019learning,chen2019graph,li2021representing}. 
Motivated by these characteristics, we first {model the spatial relation between nearby objects by organizing them as graphs and extracting latent representations through graph convolution. 
Then, we adopt the LSTM-based method to model the temporal evolution of such object relations. 
We therefore model the spatiotemporal relations among objects across space and time. }

\bfsection{Graph definition} For each frame, the nodes in graph $G_j$ correspond to a set of objects $X_j=\{x_{j,i}|i=1,\dots,n_j\}$, where $j\in \{$car, pedestrian, traffic$\}$, $\sum n_j = N$ is the total number of top identified objects. (To be convenient, we ignore frame $t$ here.) 
We construct graph $G_j \in \mathbb{R}^{n_j \times n_j}$ to represent the pairwise relations among objects in $j$-category, where relation value of each edge measures the interactions between two objects. Currently, we assume each $G_j$ is a complete graph with all edges equal to 1. The importance of the edge value and the edge type will be explored in our future work. 

\begin{figure}[t]
    \centering
    \includegraphics[width=0.7\linewidth]{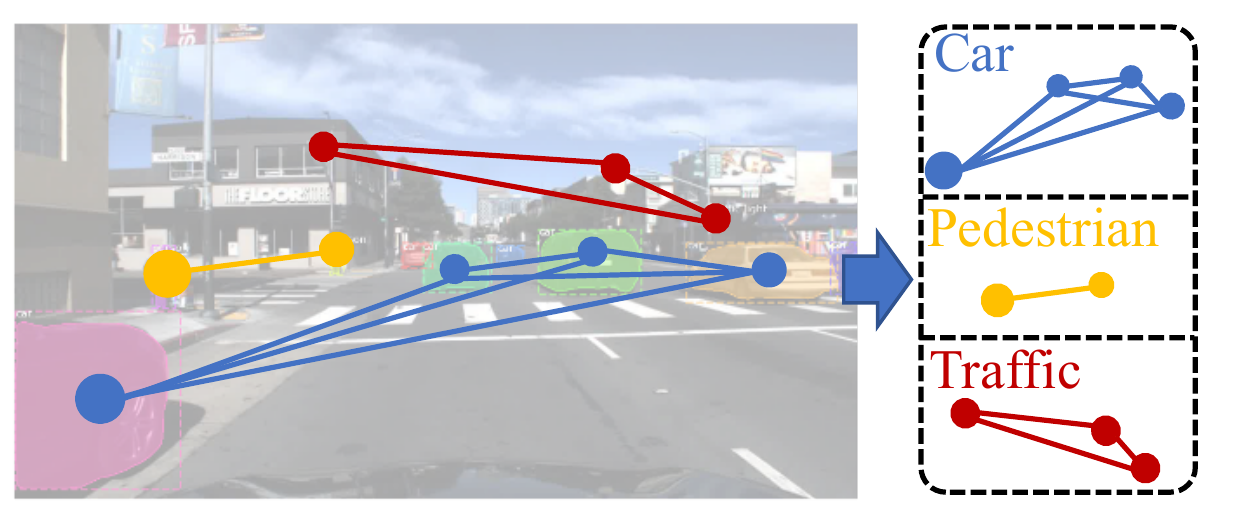}
    \caption{Illustration of building multi-view object relation graphs.}
    \label{fig:org}
\vspace{-0.15in}
\end{figure}

\bfsection{Multi-view object relation graphs} A single object relation graph $G^t_j$ typically refers to the $j$-category in a specific frame $t$. To capture dynamic patterns of objects in different category, we build a sequence of graphs $\mathcal{G}=\{G^{t}_j|t=1,\dots,T\}$ on the object feature matrices $\{X^t_j|t=1,\dots,T\}$ of objects belonging to the $j$-category across different frames, where $T$ is the number of frames. As a result, there are 3 graphs in each frame and $3 \times T$ graphs in total for each sample. Figure~\ref{fig:org} illustrates this process. Building such multi-view graphs allows us to capture the local spatial relations of objects from the same category at different frames and then jointly consider the relations of relative changes across time from multiple view. Therefore, our approach can learn the spatiotemporal relations between objects more sufficiently. 

\bfsection{Graph convolution} Different from convolution operations on images, the graph convolutions allow us to compute the response of a node based on its neighbors defined by the graph relations. To save computational cost and extract localized features, we apply the $K$-hop localized spectral graph convolution~\cite{kipf2017semi}. Specifically, if $K=1$, we only use the features of the nearest object and itself to update the representations of each object. However, the nearest object is insufficient to represent the localized features. As mentioned in Section~\ref{sec:intro}, the speed-control behavior is not only affected by the nearest object, but also by the relative position of nearby objects. To consider high-order neighborhood of each object, we can extract more sufficient information from its neighboring objects to reveal its moving pattern.

Formally, given feature matrix $X \in \mathbb{R}^{n \times 4}$ and adjacency matrix\footnote{{For frames with few objects, we make up the 0-vector in addition to identified object features and assume these objects are isolated with diagonal adjacency matrix.}} $A = 1 \in \mathbb{R}^{n \times n}$, we compute the normalized Laplacian matrix $L=I - D^{-\frac{1}{2}}AD^{-\frac{1}{2}}$,where $D=diag(A)\in \mathbb{R}^{n \times n}$ is the degree matrix. (To be convenient, we ignore the category $j$ and frame $t$ at this part.) 
The Laplacian matrix $L$ is symmetric positive semi-definite such that it can be diagonalized via eigen-decomposition as $L=U\Lambda U^T$,where $\Lambda$ is a diagonal matrix containing the eigenvalues, $U$ consists of the eigenvectors, and $U^T$ is the transpose of $U$. 
We can approximate the spectral convolution by employing with $K$-th order Chebyshev polynomial,
\begin{equation}
    G^{'}_K(X) = U \sum_{j=0}^{K}\theta_j^{'}\Lambda^{j} U^T X = \sum_{j=0}^{K}\theta_j^{'} L^{'} X.
    \label{gcn-formular2}
\end{equation}

Therefore, it is much more efficient with only $K$ learnable parameters $\theta^{'} \in \mathbb{R}^K$ and no more eigen-decomposition of $L$. Besides, it also updates features of each object node by aggregating $K$-hop neighborhood information.

\subsection{Modeling Temporal Dependencies} 
Temporal context information of each frame is crucial to capture the dynamic dependencies of objects in the consecutive scenes. Particularly, for long-range temporal dependencies, it is necessary to learn the semantic meaning of relative relations 
from the movements of objects in the scenes. In order to model the dynamics of both shape and location changes, we first max pool the output features of graph convolution over nodes to aggregate the local spatial relation information and then apply a standard multi-layer LSTM to extract the temporal features of objects from the same category. 
Because the movements of objects from different categories are significantly different, we pay attention to the different underlying patterns of moving objects in the scene.
For example, the objects from the traffic-category are always static, while the pedestrians and cars are moving randomly based on their own intentions. To implement this idea, we propose \textbf{parallel} LSTMs with joint training. Specifically, we jointly train three sub-LSTMs on the training set. Each of the sub-LSTMs models the moving pattern of objects from the same category. We parallelize the three sub-LSTMs and merge them with a concatenate layer to jointly infer the speed-control behavior.  

Finally, we can predict the speed-control behavior of a driver as follows:
\begin{equation}
    \begin{aligned}
    H_j & = \mathrm{MaxPool}(G^{'}_K(X)) \\
    E_j & = \mathrm{LSTM_j}(H_j) \\
    E &= \mathrm{Concat}(E_{\mathrm{car}},E_{\mathrm{pedestrian}},E_{\mathrm{traffic}}) \\
    \hat{y} &= \sigma (W_T E + b_T).
    \end{aligned}
\end{equation}
where $H_j \in \mathbb{R}^{T \times d}$ is the aggregated representation of the object relation graph in $j$-category, $j \in \{$car, pedestrian, traffic$\}$. $E_j \in \mathbb{R}^e$ is the spatiotemporal features extracted from the object relation graph in $j$-category. $E \in \mathbb{R}^{3\times e}$ is the final feature. $W_T$ and $b_T$ are weight matrix and bias vector, respectively. $\sigma(.)$ denotes an activation function, and we adopt Softmax here.

Overall, the EgoSpeed-Net can be trained in an end-to-end manner. We adopt the Cross Entropy loss for this multi-class prediction problem.
\begin{equation}
    \mathrm{\mathcal{L}} =  -\sum_{i=1}^{N} \sum_{j=1}^{4} {y_{i,j}\log \hat{y}_{i,j}},
\end{equation}
where $y_{i,j}$ is the ground-truth behavior label and $\hat{y}_{i,j}$ is the softmax probability for the $j$-th class of sample $i$, respectively. $N$ is the total number of egocentric video clip samples in a training batch. 
\section{Experiments}
\label{sec:evaluation}
In this section, we first introduce the dataset and the implementation details of our proposed EgoSpeed-Net. Then, we compare the performance with the state-of-the-art methods. We also conduct extensive experiments to validate the effectiveness of proposed components in our model. Finally, we demonstrate a case study.

\bfsection{Dataset} We adopt the Honda Research Institute Driving Dataset (HDD)~\cite{ramanishka2018toward}. 
The dataset includes 104 hours of real human driving in the San Francisco Bay Area collected using an instrumented vehicle equipped with different sensors. The current data collection spans from February 2017 to October 2017. The video is converted to a resolution of 1280 $\times$ 720 at 30 fps. 
We down sample the egocentric video data (\ie only 3 frames per second are taken) and exclude the initial stop part (The driver in the first frame of a valid clip should be moving). Besides, we also exclude vague video clips that are difficult to identify objects in the scene, including heavy rainy road condition, overexposed light condition, dark night condition, etc. 
At current stage, we aim to forecast driver behavior on both highway and urban scenarios without turning (\ie steering wheel angle is within range [-30, 30] degrees). Based on the observations, drivers mostly keep driving forward except for changing routes or overtaking. 
In reality, if a driver wants to change the direction, he/she has to observe the surroundings through the rear-view mirror for safety. Therefore, such decisions are self-determined and it is hard to predict without information from the rear mirror. We leave it for future work. 
After processing the data as described above, we generated 58721 video clips in total (29275 from highway while 29446 from urban) and we use 70\% (41105 samples) for training, 10\% (5872 samples) for validation, and rest 20\% (11744 samples) for testing. 

\begin{figure}[t!]
	\centering
	 \medskip
	\begin{subfigure}[b]{1\linewidth}
    \centering
    \includegraphics[width=.9\linewidth]{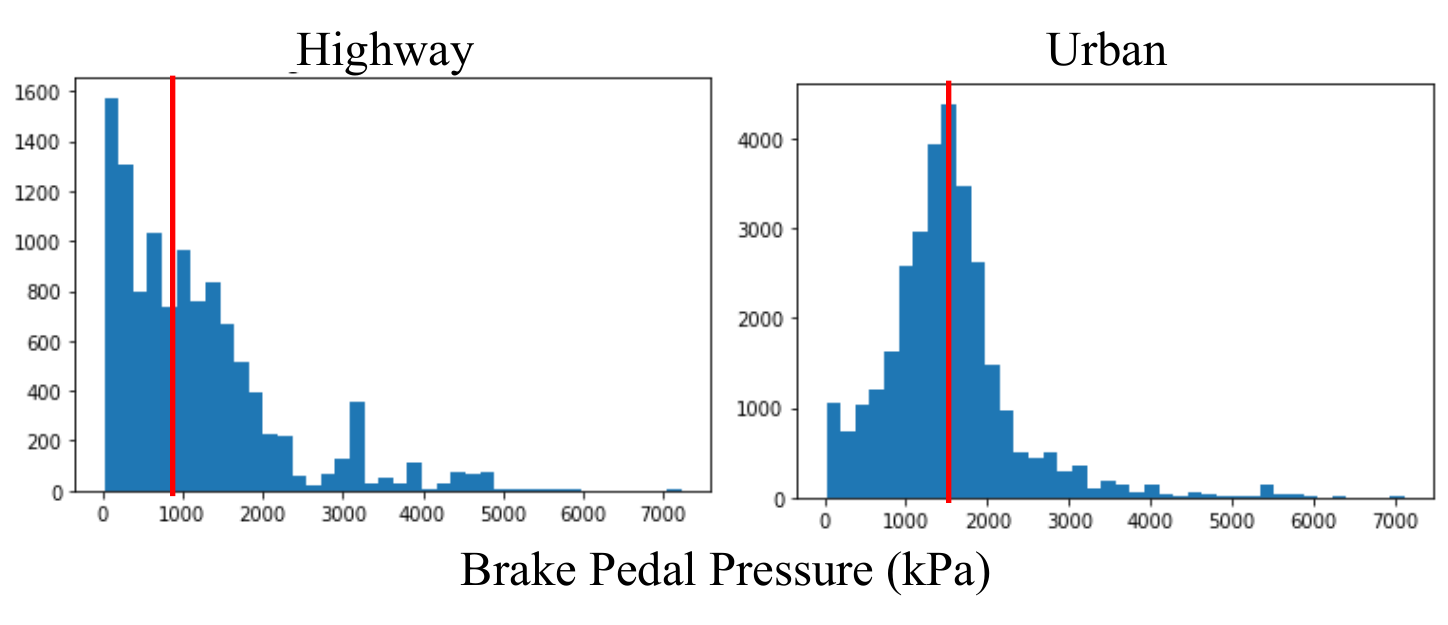}
    \caption{Histogram of braking behavior}
    \label{fig:hist_braking}
    \end{subfigure}
    
    \begin{subfigure}[b]{1\linewidth}
    \centering
    \includegraphics[width=.9\linewidth]{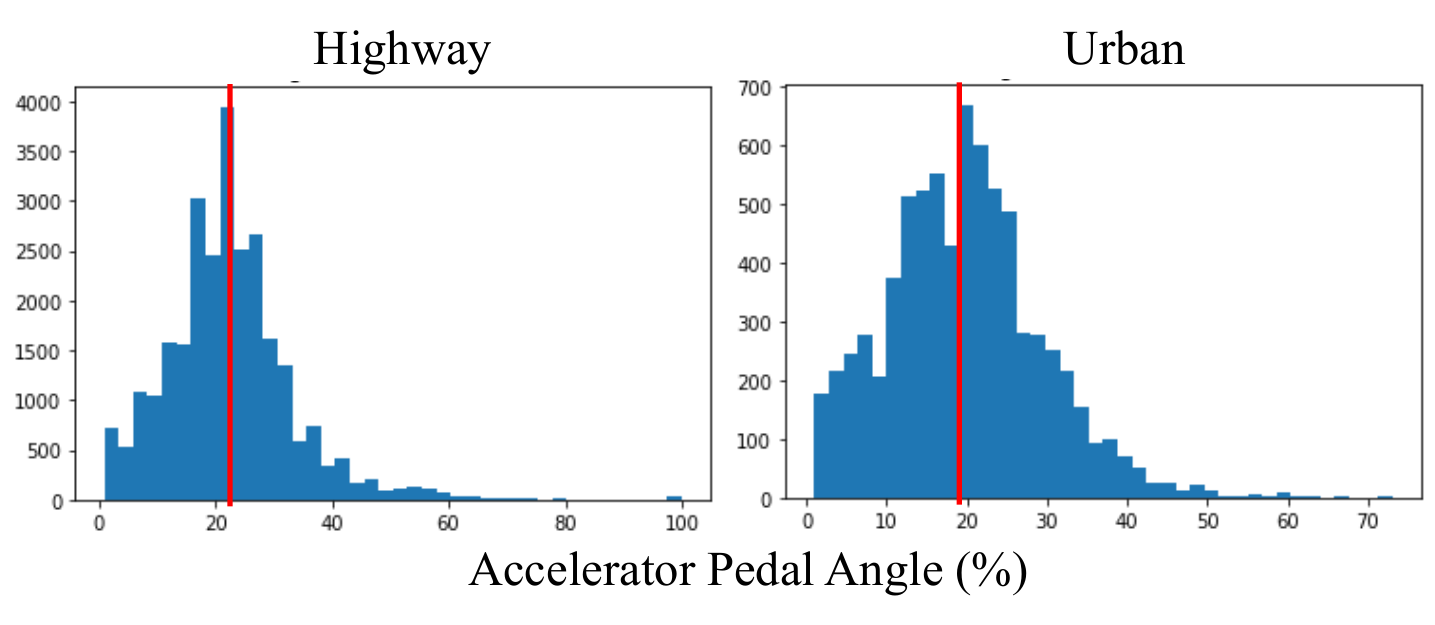}
    \caption{Histogram of acceleration behavior}
    \label{fig:hist_accel}
    \end{subfigure}
	\caption{Histogram of braking/acceleration behavior in highway and urban scenarios, respectively.}
	\label{fig:hist_behavior}
\vspace{-0.2in}
\end{figure}

\begin{table*}[t]
\caption{Comprehensive Results on EgoSpeed-Net with the State-of-the-Art Methods in terms of Recall and Accuracy (\%) (Settings: $T$=10, $FT$=1, and $K$=1).}
\label{tab:all}
\begin{tabular}{@{}ccccccccc@{}}
\toprule
\textbf{Method} & \textbf{\begin{tabular}[c]{@{}c@{}}Full \\ Braking\end{tabular}} & \textbf{\begin{tabular}[c]{@{}c@{}}Slight \\ Braking\end{tabular}} & \textbf{\begin{tabular}[c]{@{}c@{}}Slight \\ Acceleration \end{tabular}} & \textbf{\begin{tabular}[c]{@{}c@{}}Full \\ Acceleration\end{tabular}} & \textbf{Overall} & \textbf{GCN}   & \textbf{\begin{tabular}[c]{@{}c@{}}Multi-view \\ Graphs \end{tabular}}   & \textbf{\begin{tabular}[c]{@{}c@{}}Temporal \\ Module\end{tabular}} \\ 
\midrule
TRN~\cite{xu2019temporal} & 77.9 & 75.40  & 53.45 & 45.65& 67.00  & - & - &  \checkmark  \\
STIN~\cite{materzynska2020something} & 81.18 & 67.89 & 48.26  & 54.67 & 66.62  & - & \checkmark & \checkmark  \\
PointNet~\cite{qi2017pointnet} & 53.64 & 57.06 & 10.32  & 36.63 & 44.38 & -  & - & - \\
PointNet~\cite{qi2017pointnet}+T & 61.44 & 56.57 & 20.74  & 42.31  & 49.63  & - & - &  \checkmark  \\
\midrule
Base & 67.09 & 64.35 & 32.56  & 45.29  & 56.45 & \checkmark  & - & -  \\
Base+Single & 74.34 & 56.52 & 29.39 & 49.12 & 56.63 & \checkmark& - & - \\
Base+Multi & 79.09 & 71.96 & 37.12 & 53.90 & 65.22  &  \checkmark&    \checkmark & - \\
Base+T & 86.95& 85.57 & 54.85 & 68.43 & 77.75 & \checkmark & - & \checkmark \\
EgoSpeed-Net & \textbf{91.71} & \textbf{87.23} & \textbf{60.45} &  \textbf{79.03} & \textbf{82.78} &  \checkmark &    \checkmark   &       \checkmark  \\ 
\bottomrule
\end{tabular}%
\end{table*}

{\em Speed-control actions:} HDD~\cite{ramanishka2018toward} provides a list of vehicle's various sensors data. As shown in Figure~\ref{fig:hist_behavior}, we use the brake pedal pressure and the percentage of accelerator pedal angle to derive our driver speed-control behaviors and use the steering wheel angle to select samples without turning. 
These sensor data are only used to generate ground truth labels and not used as the input of our model.
In this paper, we define two levels of braking and acceleration behavior based on the braking pedal pressure and the percentage of accelerator pedal angle collected, respectively. We name the four actions {\em Full Braking, Slight Braking, Slight Acceleration,} and {\em Full Acceleration}. 
According to the histogram of brake pedal pressure in Figure~\ref{fig:hist_braking}, we see that the braking behavior of drivers in different environment shows different patterns. 
Therefore, we set 958 kPa and 1461 kPa (median value of its histogram) as the threshold of two braking levels for highway and urban scenarios, respectively. In other words, given a pedal pressure with 1000 kPa, we assume the driver is full braking in the highway scene, but slight braking in the urban scene. Similarly, according to the percentage of accelerator pedal angle in Figure~\ref{fig:hist_accel}, we set 22\% and 19\% (median value of its histogram) as the threshold of two acceleration levels for highway and urban scenarios, respectively.
As a result, we have 19727, 18546, 9607, and 10841 samples for full braking, slight braking, slight acceleration, and full acceleration, respectively. In the training phase, we oversample the slight acceleration and full acceleration behaviors to overcome the imbalanced data problem.
\begin{figure}[t!]
    \centering
    \includegraphics[width=.8\linewidth]{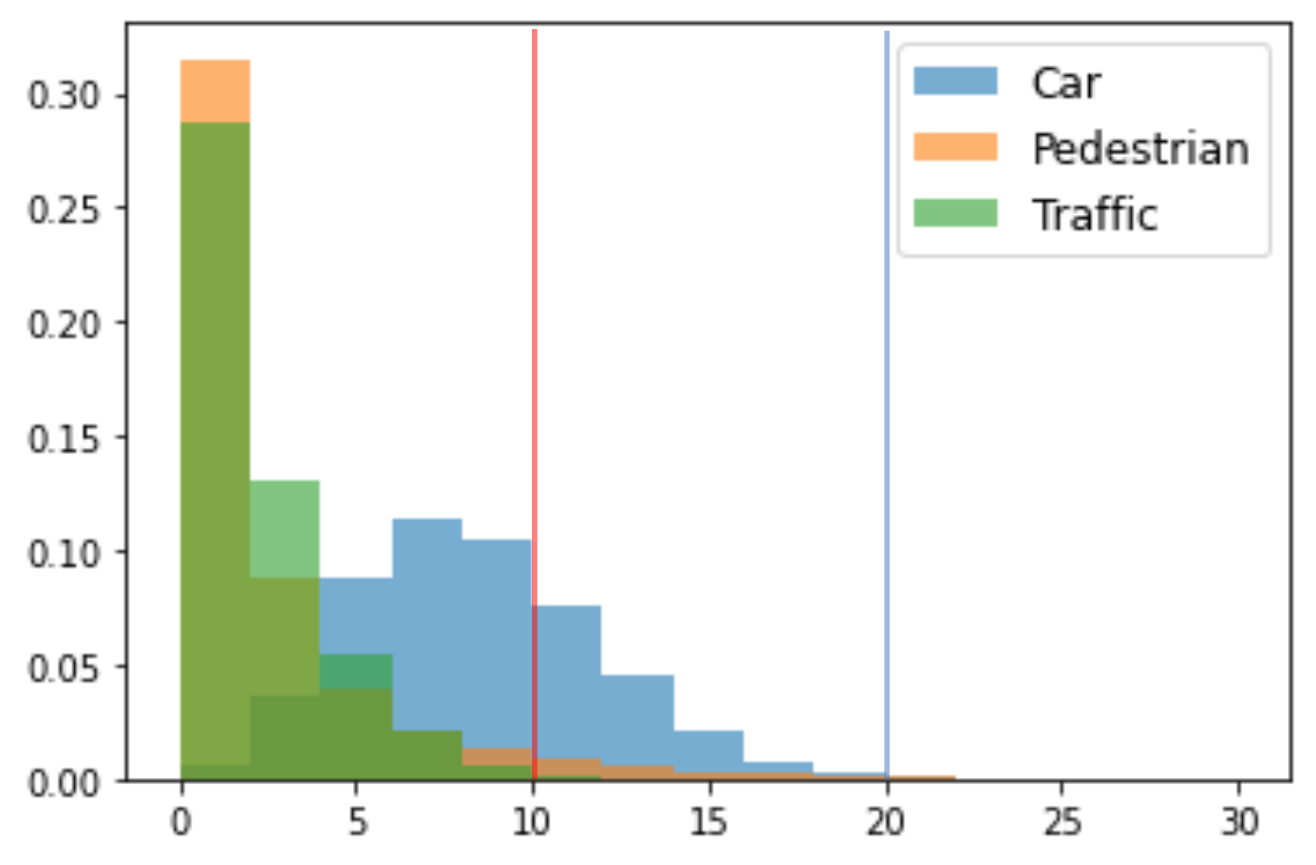}
    \caption{Histograms of identified objects from different categories.}
    \label{fig:obj_stats}
\vspace{-0.15in}
\end{figure}

\bfsection{Implementation} We set up the experiments on a High Performance Computing Cluster 
using a 256 GB RAM computing node with 2.6 GHz 16-Core CPU with Nvidia Tesla P100 Accelerator Cards. The primary development package is based on PyTorch 0.4.0 in Python 3.6.

Given a video clip with history length $T$ frame, we first apply the Mask R-CNN~\cite{lin2014microsoft} using a ResNet-50-FPN backbone with pre-trained weights on the COCO dataset
to select Top $N (=n_{\textit{car}}+ n_{\textit{pedestrian}}+n_{\textit{traffic}})$ objects in each frame according to their confidence scores in related categories. For scenes with few identified objects, we fill the object features with 0 if there are not enough objects found in the scene.
The histogram of identified objects from different categories are shown in Figure~\ref{fig:obj_stats}. To include most objects in the scene, we set $n_{\textit{car}}=20$ and $n_{\textit{pedestrian}}=n_{\textit{traffic}}=10$.
Therefore, we generate $T$ feature matrices $X$ in $N \times 4$ dimension, where $N=\sum n_j=40$ and 4 represents for the bounding box coordinates of each object in the frame. 
Then, we build object relation graphs $G$ and perform a two layer $K$-hop spectral graph convolution on them with 16 and 32 hidden nodes, respectively. 
After max pooling, the spatial feature vector is in $d$=32 dimension. We apply a two layer LSTMs with 64 hidden units and the dimension of the output spatial-temporal feature is $e$=64.
Finally, the behavior classifier consists of 2 hidden layers with 64 and 32 hidden nodes and followed by an output layer with an activation function softmax. In all experiments, we adopt early stopping criteria, set the batch size as 512, and select the Adam algorithm with default settings as our optimizer. In our experimental settings, if the validation loss decreased less than $10^{-6}$ after 50 epochs, the training will be terminated. In the end, we save the best weights instead of the latest weights.

\subsection{State-of-the-art Comparison}
\label{subsec:baseline}
To validate the effectiveness of our proposed method on the driver speed-control behaviors prediction problem, we compare the EgoSpeed-Net with the state-of-the-art methods on the Honda Dataset~\cite{ramanishka2018toward}. 
For all experiments, our problem settings include history length $T \in \{2,5,10,15\}$, future time $FT \in \{1, 5, 10\}$, and neighborhood size $K \in \{1,3,5\}$. The default setting is $T$=10, $FT$=1, and $K$=1. Here, future time indicates that instead of predicting the speed-control behavior at next frame $t+1$, we only predict he behavior at exactly the frame $t+FT$. 
Table~\ref{tab:all} summarizes the comprehensive experimental results on EgoSpeed-Net with the state of the art. 
Baseline models are as follows.
\begin{itemize}[nosep, leftmargin=*]
    \item \textbf{PointNet}~\cite{qi2017pointnet} is a well-known deep network architecture for point clouds (\ie unordered point sets) and we consider the object features $X$ as a set of point clouds. Note that we conduct the necessary normalization and argumentation on our data, including random 4d rotation and adding random noise. Because we only select Top $N=40$ objects in each frame, we set the hidden dimensions of multi-layer perceptron layers as (16,32,128,64,32) instead of (64,128,1024,512,256) in the original experimental settings.
    
    \item \textbf{Temporal Recurrent Network (TRN)}~\cite{xu2019temporal} is proposed to model greater temporal context for the online action recognition problem. For fair comparison, we exclude the parts involving sensor data and extract the same visual feature from the $Conv2d_7b_1\times1$ layer in InceptionResnet-V2~\cite{szegedy2017inception} pretrained on ImageNet~\cite{krizhevsky2012imagenet} as done in TRN. 
    
    \item \textbf{Spatial-Temporal Interaction Network (STIN)}~\cite{materzynska2020something} is proposed to model the interaction dynamics of objects composed in an action for compositional action recognition. To be fair, we use the bounding box coordinates of Top $N$ identified objects in the scene as input and follow the same hyper parameter settings in STIN. 
    
\end{itemize}

\bfsection{Metrics} Recall and accuracy are used to evaluate our model:
\begin{equation*}
\begin{aligned}
    \mathrm{Recall_j(\%)} & = \frac{n_j}{N_j} \times 100\%, j \in \{1,2,3,4\}, \\
    \mathrm{Accuracy(\%)} & = \frac{n}{N} \times 100\%, n=\sum n_j, N=\sum N_j,
\end{aligned}
\end{equation*}
{where $n_j$ and $N_j$ are the numbers of correctly predicted samples and all samples for each action class $j$, respectively. 
}

As shown in Table~\ref{tab:all}, our proposed method, EgoSpeed-Net, outperforms all the existing methods consistently with a good margin. Particularly, the Slight Acceleration behavior is the most difficult one to predict for all methods. EgoSpeed-Net is the only one whose accuracy exceed 60\%. Others are all around 50\% or less. For both Slight Braking and Full Braking behavior, EgoSpeed-Net can achieve 90\% accuracy. This outstanding performance shows the effectiveness of our approach EgoSpeed-Net with capability of simultaneously and sufficiently capturing the local spatial relationships and temporal dependencies among identified objects across space and time.

\begin{figure*}[t]
    \centering
    \includegraphics[width=.75\linewidth]{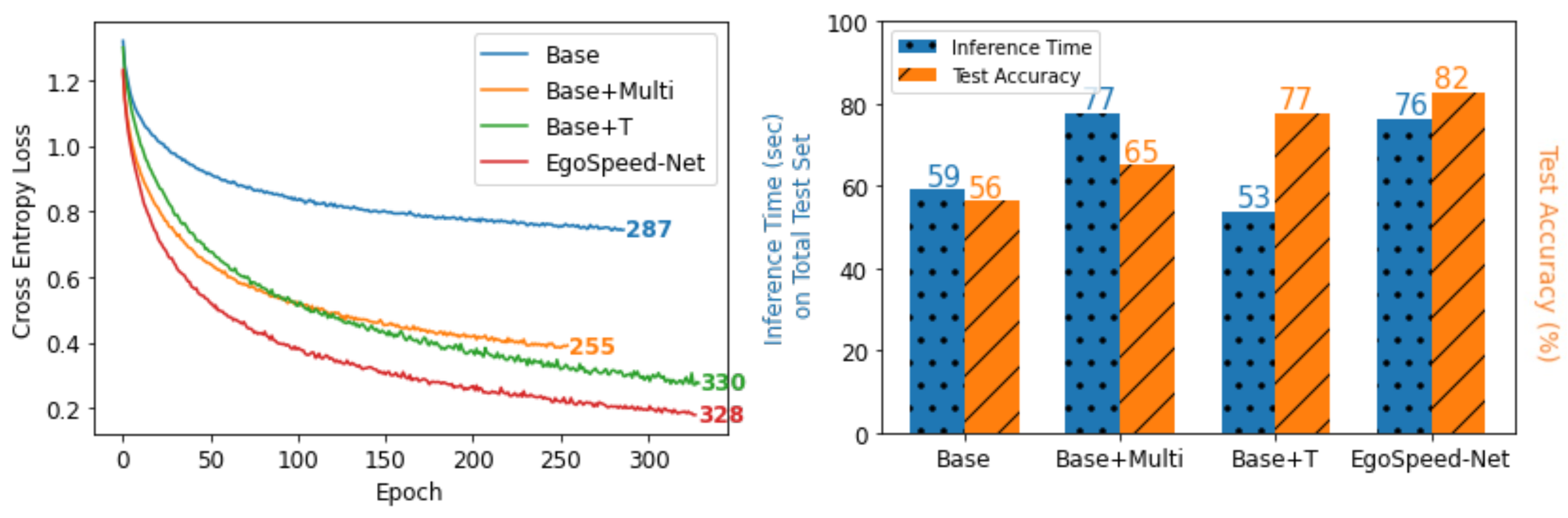}
    \caption{Comparison of training loss and inference time of different variations ($T$=10,$FT$=1 and $K$=1).}
    \label{fig:efficiency}
\end{figure*}

\subsection{Ablation Study}

In this subsection, we investigate the impact of each design in EgoSpeed-Net. 
Comprehensive results are summarized in Table~\ref{tab:all}.

\bfsections{Object relation graphs} We first evaluate the benefit of building object relation graphs over using the original object features. As an alternative, we substitute the object relation graph with a PointNet~\cite{qi2017pointnet} module as described in subsection~\ref{subsec:baseline}. For each frame, we feed the Top $N$ object features into shared Multi-Layer Perceptron (MLP) and then aggregate point features by max pooling over each image. Finally, we concatenate the informative points of point cloud across the frame and feed them into the MLP followed by a softmax function. To be fair, we compare its performance with the basic version of EgoSpeed-Net, namely, ``\textbf{Base}'' in Table~\ref{tab:all}. The base model only includes car-graph from multi-view and feeds the spatial features $H$ directly to the classifier to predict the speed-control behavior of a driver, while skipping the temporal layer. As shown in Table~\ref{tab:all}, ``Base'' outperforms PointNet for all speed-control behaviors. It demonstrates that building such graph structure allow us to model the interactions between objects in the same frame  more completely and efficiently, while assuming these object features as a set of point clouds may miss some spatial relations.

\bfsections{Multi-view graphs} We also perform the ablation study on the multi-view design to validate its effectiveness. The results are list in Table~\ref{tab:all} with label "\textbf{Base+Multi}". It suggests that providing the multi view of different categories can model more diverse relations from each other and collect more sufficient information for better representation. The accuracy from multiple views is increased by 15\% of the Base model.
In addition, we also consider building a single objects relation graph using all objects from all categories, namely \textbf{Base+Single}. Given that we observe results similar to the Base model, we conclude that simply involving more objects into the car-graph does not provide additional information. Therefore, it is necessary to build the object relation graphs from multiple views.

\bfsections{Temporal module} Furthermore, we investigate how the temporal module influences the overall performance. First, we evaluate the impact of introducing a standard two layers LSTMs into the Base model. As shown in Table~\ref{tab:all}, we observe that adding such a temporal module leads to significant and consistent improvement (overall accuracy increase up to 40\% of original value) compared with the Base model and is able to further boost overall accuracy from 77.75\% to 82\% by combining with aforementioned multi-view design. Then, we compare the performance of including an extra temporal module on other methods. For example, we add the same LSTMs layer before the speed-control behavior classifier in PointNet. We report the accuracy measures of these two methods with label "\textbf{Base+T}" and "\textbf{PointNet~\cite{qi2017pointnet}+T}" in Table~\ref{tab:all}. We see that the temporal module is helpful to improving the performance of both methods. Moreover, our Base model can take full advantage of the extracted temporal features with higher improvement compared with PointNet. Note that combining the multi-view design and temporal module design together is our proposed approach, EgoSpeed-Net, which demonstrates that our framework can effectively and sufficiently model the spatiotemporal relation between objects across space and time.

\bfsections{Efficiency analysis}
We evaluate the efficiency of each component in EgoSpeed-Net. Figure~\ref{fig:efficiency} shows the comparison of training loss curve of different variations, as well as inference time on the total test dataset. Because of the early stopping criteria, Base and Base+Multi stop before overfitting. These curves clearly show that EgoSpeed-Net learns the driver behavior efficiently at first few epochs while its loss drops significantly. Comparing the inference time of Base+Multi and Base+T, we see that the multi-view component is time consuming. Particularly, Base+T model is the most efficient one at the expense of 5\% decrease of accuracy compared with EgoSpeed-Net. However, if we average the inference time on the entire test set, the increased time of EgoSpeed-Net is still acceptable.

\begin{table*}[t]
\caption{Comprehensive Results on EgoSpeed-Net with Exploration of Different Settings ($T\in \{2,15\}$, $FT \in \{1,10\}$, and $K\in \{1,5\}$).}
\label{tab:settings}
\begin{tabular}{c|ccccccccc}
\toprule
\multicolumn{1}{l}{\textbf{Set}} & \multicolumn{1}{c}{\textbf{Method}} &\textbf{\begin{tabular}[c]{@{}c@{}}Full \\ Braking\end{tabular}} & \textbf{\begin{tabular}[c]{@{}c@{}}Slight \\ Braking\end{tabular}} & \textbf{\begin{tabular}[c]{@{}c@{}}Slight \\ Acceleration \end{tabular}} & \textbf{\begin{tabular}[c]{@{}c@{}}Full \\ Acceleration\end{tabular}} & \multicolumn{1}{l}{\textbf{Overall}} & \multicolumn{1}{l}{\textbf{\begin{tabular}[c]{@{}c@{}}History \\ Length\end{tabular}}} & \multicolumn{1}{l}{\textbf{\begin{tabular}[c]{@{}c@{}}Neighborhood  \\ Size\end{tabular}}} & \multicolumn{1}{l}{\textbf{\begin{tabular}[c]{@{}c@{}}Future \\ Time\end{tabular}}} \\
\midrule
\multirow{4}{*}{1} & Base & 65.63  & 54.43  & 20.81  & 48.28 & 51.49 & \multirow{4}{*}{2} & \multirow{4}{*}{5} & \multirow{4}{*}{1} \\
 & Base+Multi  & 74.59 & 70.20  & 25.81  & 61.90   & 62.81  &  &  & \\
& Base+T  & 68.26 & 60.02  & 24.45 & 53.48  & 55.7  &  & &  \\
& EgoSpeed-Net & \textbf{78.18} & \textbf{71.58} & \textbf{27.32} & \textbf{61.24} & \textbf{64.57}&  &   &   \\
\hline
\multirow{4}{*}{2} & Base  & 71.01 & 63.12 & 33.75 & 46.77 & 57.85 & \multirow{4}{*}{15} & \multirow{4}{*}{5} & \multirow{4}{*}{1} \\
 & Base+Multi & 77.16  & 78.53  & 37.22 & 60.22 & 67.83 & &   &  \\
 & Base+T  & 91.73 & 87.73  & 71.54 & 74.83  & 83.97   &   &   & \\
 & EgoSpeed-Net & \textbf{92.50} & \textbf{89.06} & \textbf{71.12} & \textbf{83.67} & \textbf{86.25} & &  &    \\
 \hline
 \multirow{4}{*}{3} & Base  & 71.81 & 60.18 & 24.76  & 56.77 & 57.60 & \multirow{4}{*}{15} & \multirow{4}{*}{1}  &  \multirow{4}{*}{1}\\
 & Base+Multi & 78.2 & 77.89 & 38.39 & 58.54 & 67.86  &   &  &   \\
 & Base+T & 90.36  & 87.22 & 67.16 & 80.69 & 83.74 & &  & \\
& EgoSpeed-Net  & \textbf{92.26} & \textbf{88.86} & \textbf{68.76} & \textbf{83.54} & \textbf{85.69} &  &  &  \\
\hline
\multirow{4}{*}{4}  & Base & 72.81 & 66.11 & 26.10 & 44.06  & 57.74 & \multirow{4}{*}{15} & \multirow{4}{*}{5} & \multirow{4}{*}{10}  \\
 & Base+Multi & 82.07 & 76.08 & 31.85 & 61.18 & 68.18 &  &  &  \\
 & Base+T & 91.36 & 88.87  & 72.26  & 79.89 & 85.32  &   &   &   \\
 & EgoSpeed-Net & \textbf{93.27} & \textbf{91.22} & \textbf{78.94} & \textbf{82.49} & \textbf{88.26}  &  &  &  \\                              
\bottomrule
\end{tabular}%
\end{table*}

\bfsections{Exploration of different settings} With all the design choices set, we also evaluate their performance with different settings. The results are summarized in Table~\ref{tab:settings}. Here, we list four sets of experiments. Set 2 has the basic settings with history length $T=15$, neighborhood size $K=5$, and future time $FT=1$. The variation of history length, neighborhood size and future time is shown in Set 1 ($T=2$), Set 3 ($K=1$) and Set 4 ($FT=10$), respectively. 
Comparing Set 1 and Set 2, we observe that the multi view design makes the most contribution to the improvement when history length $T=2$. As history length increases, the temporal module becomes the main contributor with 56\% accuracy improvement, while multi view only boosts 27\%. This observation shows that the longer the history, the more effective the time module. 
Comparing Set 2 and Set 3, except slight acceleration, when more objects are included in the neighborhood, the improvement of other 3 behaviors is limited. We see that the information of the nearest object is sufficient enough to capture the underlying spatial relation between objects in the scene for full braking, slight braking and full acceleration behaviors. However, the slight acceleration is the most difficult behavior to predict, which requires more information from nearby objects in the scene. 
Comparing Set 2 and Set 4, we see that our approach can predict the speed-control behavior more accurately at future frame. Given a video clip, a driver needs time to response to the environment. This reaction time differs from person to person. Therefore, it is hard to tell the most accurate predicted frame. Our model yields the best performance of slight acceleration behavior with a significant improvement of 11\%.

\begin{figure}[t]
\includegraphics[width=.9\linewidth]{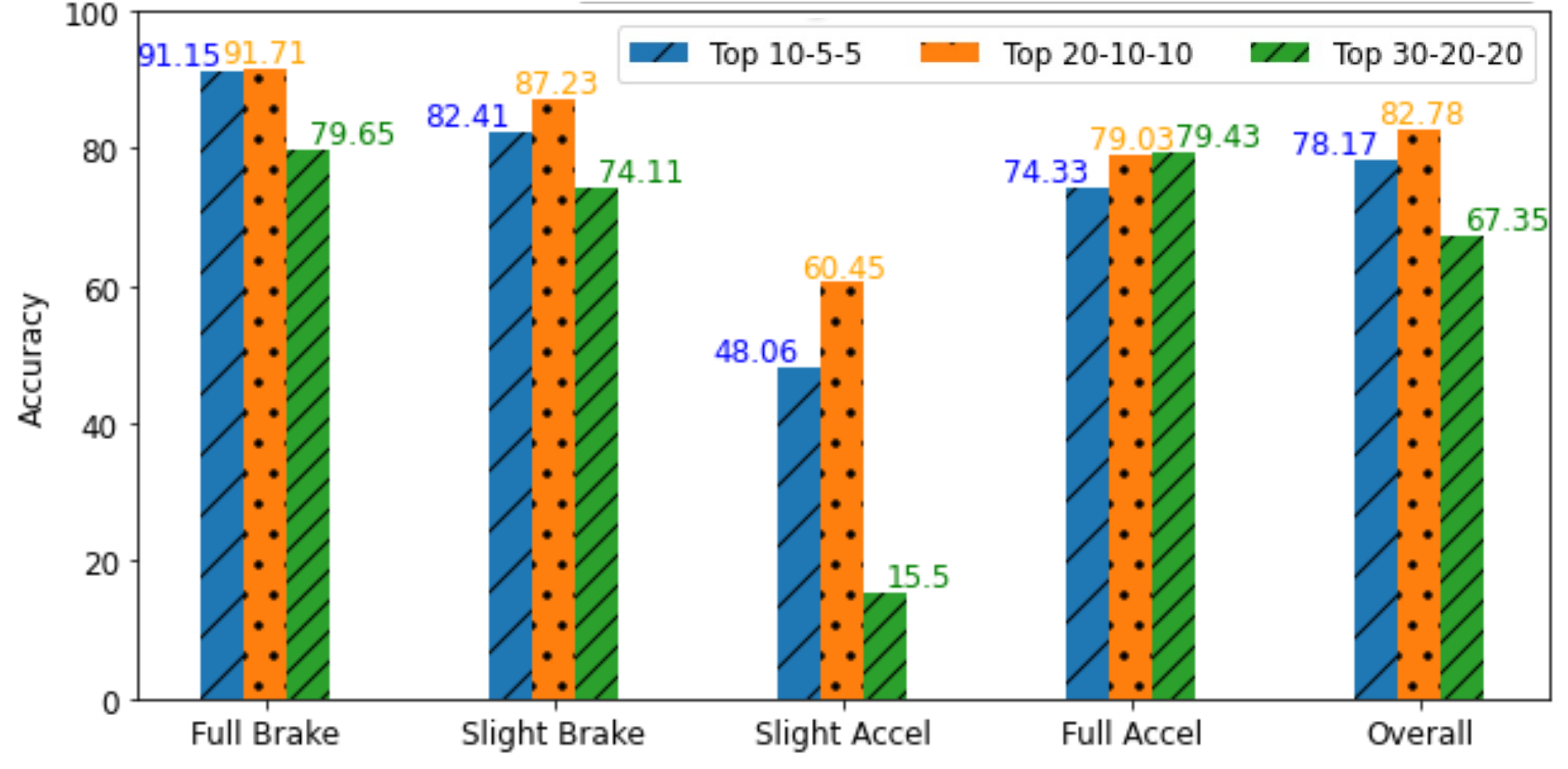}
\caption{Exploration of selecting Top $n_{\textit{car}}-n_{\textit{pedestrian}}-n_{\textit{traffic}}$ identified objects in the scene ($T=10, FT=1$, and $K=1$).}
\label{fig:abl_topn}
\end{figure}

\bfsections{Exploration of different Top $N$ objects} We also evaluate EgoSpeed-Net with different Top $N$ objects. According to the histogram of identified objects in Figure~\ref{fig:obj_stats}, we select the Top $N$ with a set of smaller numbers ($n_{\textit{car}}=10$ and $n_{\textit{pedestrian}}=n_{\textit{traffic}}=5$) and a set of larger numbers ($n_{\textit{car}}=30$ and $n_{\textit{pedestrian}}=n_{\textit{traffic}}=20$). Results are shown in Figure~\ref{fig:abl_topn}. 
According to the histogram, Top 10-5-5 select only part of objects in the scene, while Top 20-10-10 include most objects. We see that including as many objects as possible will complement the information of the scene and improve the prediction accuracy of the speed-control behaviors. However, if we select all objects in the scene with Top 30-20-20 settings, the performance will decrease. 
For example, the number of identified objects from car-related categories in most scenes is less than 20 referring to Figure~\ref{fig:obj_stats}. Therefore, changing the number to a larger number (e.g., 30) may not make much difference and may cause the matrix to be very sparse. Similar to the pedestrian and traffic-related categories, selecting too many objects may lead to the sparse matrix. 
As a result, we select the best settings with Top $n_{\textit{car}}=20$ and $n_{\textit{pedestrian}}=n_{\textit{traffic}}=10$ in our experiments.

\begin{figure*}[t!]
	\centering
	 \medskip
	\begin{subfigure}[b]{1\linewidth}
    \centering
    \includegraphics[width=.9\linewidth]{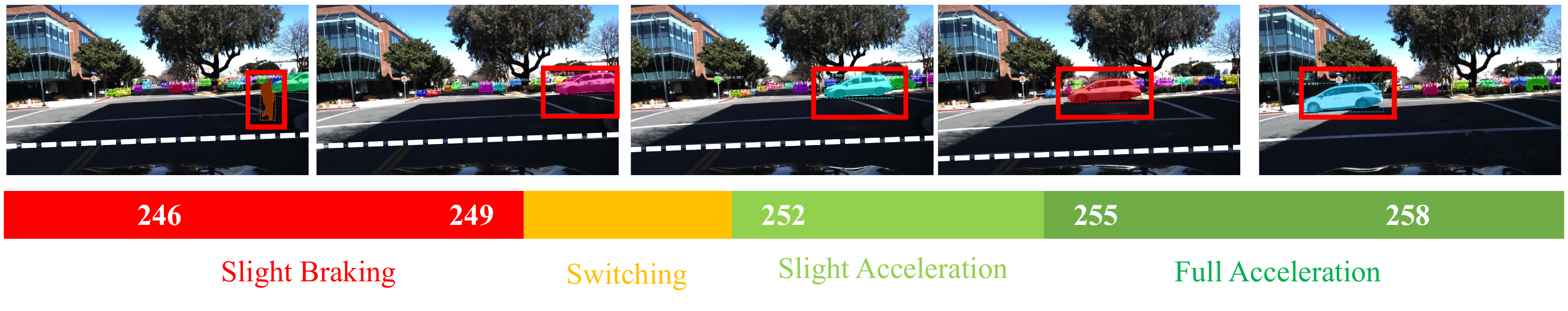}
    \vspace{-0.1in}
    \caption{Slight Acceleration}
    \label{fig:demo-a}
    \end{subfigure}
    
    \begin{subfigure}[b]{1\linewidth}
    \centering
    \includegraphics[width=.9\linewidth]{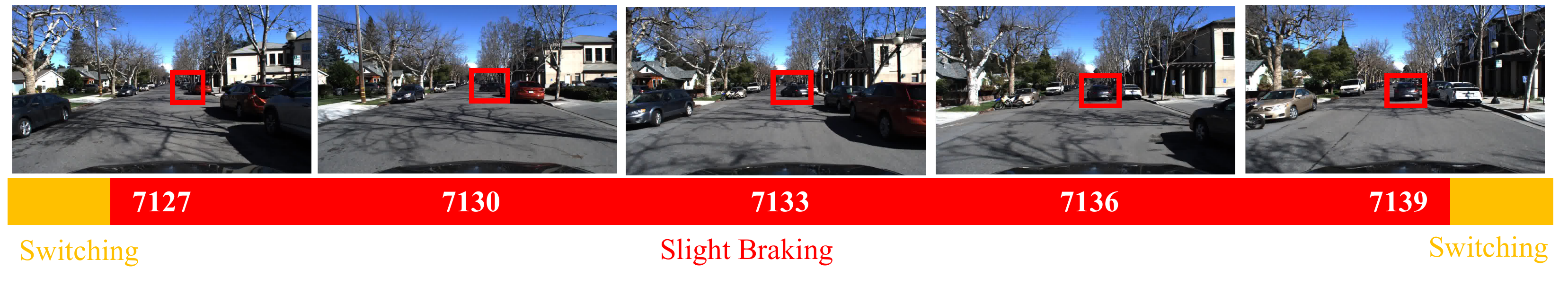}
    \vspace{-0.1in}
    \caption{Slight Braking}
    \label{fig:demo-b}
    \end{subfigure}
	\vspace{-0.2in}
	\caption{Demonstration of the slight acceleration and braking behavior.Different colors represent different speed-control behaviors.}
	\label{fig:demo}
\end{figure*}

\subsection{Case Study}
As shown in Table~\ref{tab:all} and~\ref{tab:settings}, the slight acceleration is the most difficult behavior to predict with the lowest accuracy among the speed-control behaviors. Figure~\ref{fig:demo-a} showcases the speed-control behaviors of a driver during a few consecutive frames (245-259). To be convenient, we sample one frame per second to represent the scene observed along the route of the driver. After the pedestrian has crossed the road, the driver has already started to accelerate even if there is a car moving horizontally from right to left in the scene. In detail, the driver starts accelerating at frame 252 and there is a certain distance from the crosswalk. After slightly accelerating for 3 frames, the driver approaches the crosswalk and starts fully accelerating. Comparing frame 249 and frame 252, except for the moving car marked in the red rectangle, there is no changes of other objects in the scene. Based on the relative change of the locations of this moving car, the driver decides to accelerate. Compared with other methods, our model can predict it successfully. This example demonstrates that our approach can capture such small relative changes of objects in few consecutive frames and reveal the underlying pattern of slight acceleration behavior. Another example in Figure~\ref{fig:demo-b} shows the slight braking behavior from frame 7127 to 7139. When a car appears from a cross, the driver decides to slight braking. At frame 7133, our model predicts it as full braking, which is acceptable. We see that it is hard to distinguish between slight braking and full braking in lack of the speeding information. In future, we will extract more information from the scene to complement the video understanding.

\section{Conclusion}
\label{sec:conclusion}
This paper investigated the problem of predicting the speed-control in driver behavior. Given a segment of egocentric video recorded from a continuous trip, we aim at learning a model to predict the speed-control behavior of a driver based on the visual contents from his/her point of view in the past few seconds. This problem is important for understanding the behaviors of drivers in road safety research. Prior work did not address this problem as they either used static camera data or only predicted behaviors of targeted objects in the scene rather than the observer. Few of them estimated the egocentric trajectories but required extra data (\eg trajectories, or sensor data). In this paper, we proposed EgoSpeed-Net, a GCN-based framework to address the problem. EgoSpeed-Net uses multi view object graphs and paralleled LSTMs design to model diverse relations of objects in the scene across space and time. Experiment results on the HDD showed that our proposed solution outperforms the state-of-the-art methods. 

\section*{Acknowledgements}
Yichen Ding and Xun Zhou were funded partially by Safety Research using Simulation University Transportation Center (SAFER-SIM). SAFER-SIM is funded by a grant from the U.S. Department of Transportation’s University Transportation Centers Program (69A3551747131). However, the U.S. Government assumes no liability for the contents or use thereof. Ziming Zhang was supported in part by NSF grant CCF-2006738. Yanhua Li was supported in part by NSF grants IIS-1942680 (CAREER), CNS-1952085, CMMI-1831140, and DGE-2021871. 

\bibliographystyle{ACM-Reference-Format}
\bibliography{reference}

\end{document}